\pdfoutput=1

\documentclass[11pt]{article}

\usepackage{EMNLP2023}

\usepackage{times}
\usepackage{latexsym}

\usepackage[T1]{fontenc}

\usepackage[utf8]{inputenc}

\usepackage{microtype}

\usepackage{amssymb}
\usepackage{amsmath}
\usepackage{tcolorbox}
\usepackage{booktabs}
\usepackage{graphicx}
\usepackage{CJKutf8}
\usepackage{multirow}
\newcommand{\name}{\textsc{Self-QA}}

\title{\name: Unsupervised Knowledge Guided \\ Language Model Alignment}

\author{Xuanyu Zhang \textmd{and} Qing Yang \\
\\
Du Xiaoman Financial \\
}

\begin{document}
\maketitle
\begin{abstract}
Large-scale language models like ChatGPT and GPT-4 have gained attention for their impressive conversational and generative capabilities. 
However, the creation of supervised paired question-answering data for instruction tuning presents formidable challenges. This endeavor necessitates substantial human effort for data annotation and wrestles with issues concerning data quality, diversity, accuracy, and other related factors.
To overcome these obstacles, we introduce an innovative framework named \name, which replaces the traditional practice of human-written instruction seeds with a vast amount of unsupervised knowledge, enabling the model to generate a larger quantity of correct and domain-specific instruction data.
The effectiveness of our proposed method is demonstrated through experiments conducted on unsupervised corpora from various domains.

\end{abstract}

\section{Introduction}

With the emergence of GPT-based \cite{radford2018improving} large-scale models like InstructGPT \cite{ouyang2022training}, ChatGPT \cite{openaichatgpt} and GPT-4 \cite{openai2023gpt4}, their remarkable conversational and generative capabilities have garnered widespread attention. These models not only have the capacity to understand complex language structures and grasp subtle meanings but also possess the remarkable capability to interact naturally and fluently with users, generating text that is both coherent and highly creative. This has pushed the boundaries of what was previously deemed impossible. The impact of these large-scale models extends beyond the academic realm of natural language processing (NLP) and has a profound influence in the domains of business and industry. They have opened up new possibilities for human-machine interactions, intelligent customer service, and virtual assistant applications, revolutionizing these fields and paving the way for innovation and advancement.

\begin{figure}
\centering
\includegraphics[scale=0.47]{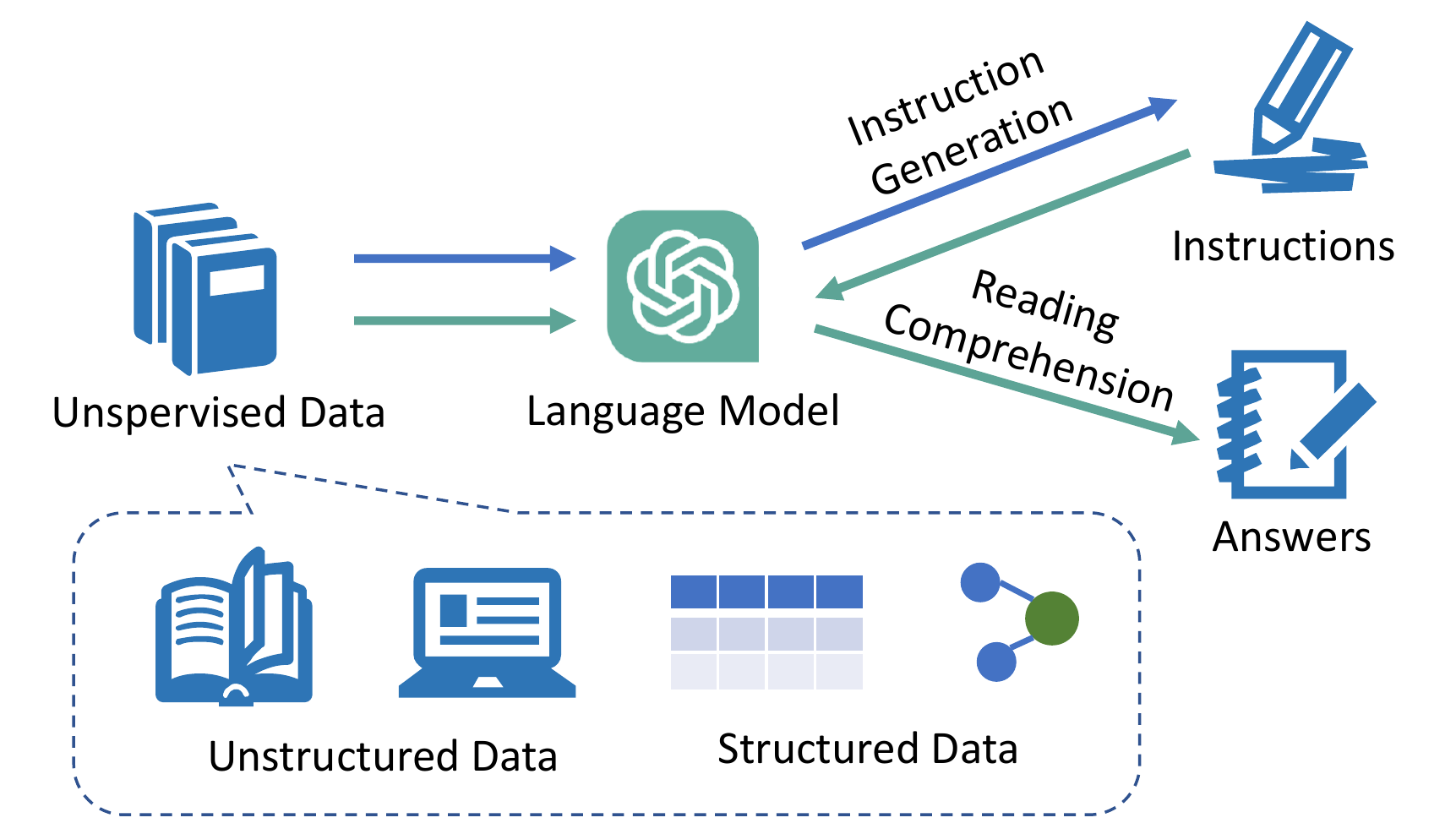}
\caption{The pipeline of \name.}
\label{fig:intro}
\end{figure}

Despite the impressive capabilities of ChatGPT, constructing supervised fine-tuning (SFT) data for instruction tuning presents significant challenges. The human effort required for annotating  data, along with issues related to data quality, diversity, accuracy, and others, hinder the development of this technique.
Although Self-Instruct \cite{wang2022self} has been proposed to mitigate this issue, it still relies on a small set of human-written seed instructions for guidance. Furthermore, the method is limited in its ability to control the domain coverage of generated instruction data and ensure the correctness of the generated answers. 
Consequently, there is a vast amount of untapped potential in utilizing the abundant unsupervised data, particularly domain-specific expertise.

\begin{table*}
\centering
\begin{tabular}{llcc}
\hline
\multirow{2}{*}{\textbf{Model}} & \multirow{2}{*}{\textbf{Prompt}} & \textbf{Domain} & \textbf{Correctness} \\
& & \textbf{customization} & \textbf{guarantee} \\
\hline
Self-Instruct \cite{wang2022self} & 176 human-written seeds & \texttimes & \texttimes \\
Self-Align \cite{sun2023principle} & 195 human-written seeds & \texttimes & \texttimes \\
Self-Chat \cite{xu2023baize} &  111,502 supervised dialogues & \texttimes & \texttimes \\
\textbf{\name \;(ours)} & \textbf{Unsupervised knowledge}  & \textbf{\checkmark} & \textbf{\checkmark} \\
\hline
\end{tabular}
\caption{Comparison of different self-alignment methods.}
\label{tab:related}
\end{table*}

Therefore, in this paper, we introduce \name, a framework to generate SFT data from unsupervised knowledge, inspired by the human self-questioning learning approach.
\name \;replaces manually written seeds used in other self-alignment models \cite{wang2022self,sun2023principle,xu2023baize} with a vast amount of unsupervised knowledge, alleviating the difficulty of language models in generating instruction data according to specific requirements.
As shown in Figure \ref{fig:intro}, 
the unsupervised data are used sequentially in the stage of knowledge-guided instruction generation  and machine reading comprehension.
\name \;not only reduces the reliance on human annotators but also allows for the generation of diverse, correct, and domain-specific instruction data.
Experiments with unsupervised corpora from various domains demonstrate the effectiveness of our proposed method.

\section{Related Work}

\paragraph{Language Models with Instruction-tuning} Recently, numerous studies \cite{ouyang2022training,peng2023instruction} have investigated the effectiveness of language models in following instructions by leveraging annotated instructional data.
This approach enables the model to learn to identify and extract relevant information from different types of instructions and use it to generate accurate and relevant responses. It enhances the model's ability to understand complex instructions and generalize to new tasks by exposing it to a wide range of instructional scenarios. However, the reliance on human annotation in creating such instructional datasets presents a bottleneck for scaling up and achieving broader applicability of instruction-guided language models. 
To address this limitation, researchers have explored alternative approaches that reduce the need for extensive human involvement in generating instruction data.

\paragraph{Bootstrapped Instruction Generation} Bootstrapped instruction generation is a recently proposed class of methods \cite{wang2022self,sun2023principle,xu2023baize} that reduces the cost of human instruction annotation.
For example, Self-Instruct \cite{wang2022self} is proposed to enhance the ability of pre-trained language models to follow instructions by utilizing their own generated samples. This technique involves generating a set of instruction, input, and output samples from the instruction seeds, and then carefully pruning them before fine-tuning the model. 
Self-Align \cite{sun2023principle} primarily employs topic-guided red-teaming self-instruct and principle-driven self-alignment to tackle the challenges associated with heavy human annotations. 
It aims to develop AI agents capable of generating helpful, ethical, and reliable responses to user queries, including adversarial ones, while proactively addressing harmful inquiries in a non-evasive manner.
However, as shown in Table \ref{tab:related}, these methods often require a small amount of supervised seed information.
The instructions generated by them cannot specify domains and content, nor can they ensure the accuracy and professionalism of the instruction responses.
Different from them, our approach can effectively address these issues by leveraging unsupervised knowledge.

\begin{figure}
\centering
\includegraphics[scale=0.8]{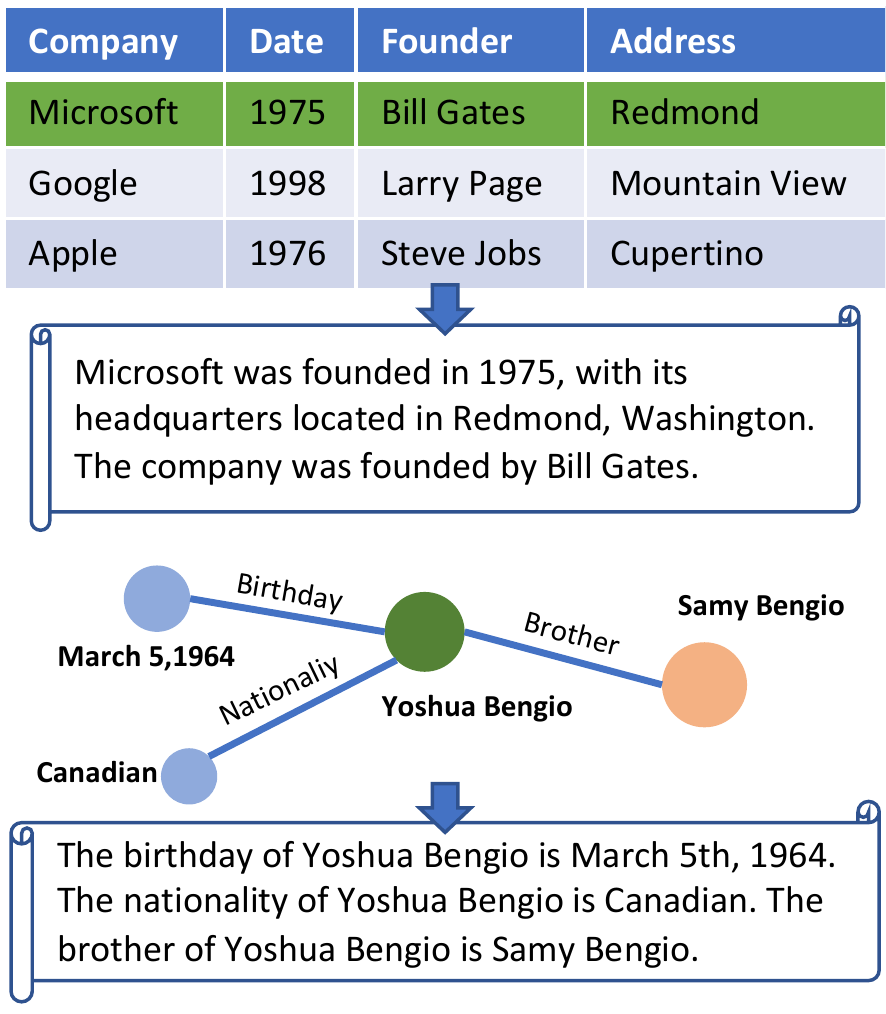}
\caption{Examples of transformation of unsupervised structured data.}
\label{fig:transform}
\end{figure}

\paragraph{Question Generation and Answering} Question generation and question answering are two closely related tasks in natural language processing. They can be viewed as a dual problem, where the former involves creating questions from a given passage or set of information, and the latter involves answering questions based on a given passage or set of information. Especially, the technique of machine reading comprehension (MRC) \cite{zhang-2019-mc,Zhang2020RceptionWA} is often used for question answering.
For humans, self-questioning and self-answering learning entail stimulating individuals to formulate their own questions and answers based on the provided information, followed by comparing their responses to the original knowledge. This approach has showcased encouraging outcomes in augmenting individuals' understanding of the provided information \cite{joseph2016effects}.
For domain-specific instruction samples, instruction and input can often be considered as a whole. Therefore, in this paper, we assume that \textbf{instructions} are equivalent to \textbf{questions}, and \textbf{instruction outputs} are equivalent to \textbf{answers}.

\section{Methodology}

Our proposed \name \;consists of three different stages: knowledge-guided instruction generation, machine reading comprehension, and filtering and pruning.

\paragraph{Knowledge-Guided Instruction Generation} In this stage, we employ the language model itself to generate instructions according to unsupervised text. This approach makes the generated instructions domain-specific and content-relevant to the unsupervised text provided. However, in the process of training and inference, instructions are fed to language models without background knowledge, so we need to provide some guidelines so that these instructions cannot rely on and refer to the content in the original text. For instance, the prompt can be:

\begin{tcolorbox}[colback=blue!10!white,colframe=blue!50!black,title=Instruction Generation Prompt]
\small
The background knowledge is:\\
\{unsupervised knowledge data\}\\

Please generate ten instruction questions as diverse as possible based on the content of the above article. These questions can be questions about facts or an understanding and evaluation of relevant content. Please assume that there is no corresponding article to refer to when asking questions, so do not use demonstrative pronouns such as ``this'' or ``these'' in the question.\\

Please generate questions in the following format:\\
1. Question: ...\\
2. Question: ...
\end{tcolorbox}

Then we can obtain several related instructions, which can be used in the next stage. \textit{\{unsupervised knowledge data\}} in the prompt represents sequential text. 
Unstructured knowledge, such as web pages and book data, can be used directly after undergoing cleaning processes. Structured data such as tables and knowledge graphs \cite{zhang-etal-2022-trans} need to be converted into unstructured textual data before they can be utilized. 
As shown in Figure \ref{fig:transform}, this can be achieved by filling slots using templates or by concatenating each data entry with its corresponding attribute name.

\paragraph{Machine Reading Comprehension} In this stage,
the language model needs to generate answers to the generated instruction questions according to the corresponding unsupervised knowledge.
The process can be formulated as follows:
\begin{equation}
\label{eq:condition}
P({\bf A}|{\bf K},{\bf Q})  = \prod_{j} P({\bf A}_i|{\bf A}_{\leq i}, {\bf K},{\bf Q})  
\end{equation}
where ${\bf k, Q, A}$  represents unsupervised knowledge, instruction question, and answer, separately. 
Because the whole process is the same as that of reading comprehension, we also call this stage by this name.
As in the previous stage, the prompt for the reading comprehension stage is as follows:

\begin{tcolorbox}[colback=blue!10!white,colframe=blue!50!black,title=Reading Comprehension Prompt]
\small
The background knowledge is:\\
\{unsupervised knowledge data\}\\
Please answer the following question based on the content of the article above:\\
\{the generated question\}\\

Please answer this question as thoroughly as possible, but do not change the key information in the original text, and do not include expressions such as ``based on the above article'' in the answer.\\

Please generate the corresponding answer in the following format:\\
Question: ...\\
Answer: ...
\end{tcolorbox}

\paragraph{Filtering and Pruning} Although we explicitly instruct the model to assume no prior knowledge from external documents and prohibit the use of demonstrative pronouns like ``this'' in generated questions and the phrase like ``based on the above content'' in generated answers,
we still observed that the language model still produces text that violates these rules.
Additionally, the generated instances of instructions also exhibit cases where they do not adhere to the required format and become unparseable.
Therefore, it is necessary to further filter out these problematic examples. 

To mitigate these issues, we implement a post-processing step to filter out inappropriate responses and correct any formatting errors. This involves developing heuristics and rule-based methods to identify and remove instances that violate the instructed constraints. By applying these filters, we ensure that the generated text adheres to the predefined guidelines and maintains the desired level of correctness and coherence.

\section{Discussion}
\subsection{Performance}

We collect several domains of unsupervised unstructured and structured data for experiments.
An example of unsupervised  knowledge and generated instruction questions and answers are shown in Table \ref{tab:example1}.
We then instruction-tuning BLOOM-7B \cite{scao2022bloom} with these generated instructions.
As shown in Table \ref{tab:example2},
our model can answer the corresponding question correctly,
but ChatGPT gives a wrong answer.
It is precisely because of these domain-specific instruction-tuning data that our model has achieved better performance.

\begin{table}
\small
\centering
\begin{tabular}{lp{4cm}}
\toprule
\textbf{Knowledge:} & Company: DXM Founding Date: April 28, 2018 Formerly known as: Baidu Financial 
Headquarters Address: Haidian District, Beijing, China.\\

\midrule
\textbf{Question1:} & When was DXM founded? \\
\textbf{Answer1:}& DXM was founded on April 28, 2018.  \\
\midrule
\textbf{Question2:} & Where is the headquarters of DXM located? \\
\textbf{Answer2:}& The headquarters of DXM is located at Haidian District, Beijing, China.  \\
\bottomrule
\end{tabular}
\caption{Examples of unsupervised background knowledge  and generated  question and answer pairs.}
\label{tab:example1}
\end{table}

\begin{table}
\small
\centering
\begin{tabular}{lp{4cm}}
\toprule
\textbf{Human:} & Where is DXM? \\ 
\midrule
\textbf{ChatGPT:} & The headquarters of DXM is located in Hangzhou, China. \\
\midrule
\textbf{Our Model:} & DXM is a financial technology company headquartered in Haidian District, Beijing, China. \\
\bottomrule
\end{tabular}
\caption{Answers of different models.}
\label{tab:example2}
\end{table}

\subsection{Different Stages of \name}
The stage of knowledge-guided instruction generation and machine reading comprehension can also be integrated into a single stage so that the model only needs to be invoked once for each round of instruction generation and answer prediction. 
The advantage of this is that the number of calls to the model is reduced, because each round of instruction question and answer generation only needs language models once.
However, there are also potential drawbacks to this approach. For instance, the model may generate output that exceeds the predetermined length. Additionally, by combining these two tasks, the model may not be able to focus on a single task as effectively, which can result in less detailed and accurate answers. Therefore, the decision to integrate two stages into a single stage should be made with careful consideration of the specific application and task requirements.

\subsection{Different Forms of Knowledge}

In general, knowledge can be stored in large language models in a parametric manner or separately input into the models in an explicit symbolic form. The main focus of this paper is on how to store unsupervised knowledge in large models using a parameterized approach. This approach enables end-to-end processing of user questions and optimization of model parameters without the need for external information. 
It offers a high level of flexibility and adaptability to different inputs and contexts.
However, this approach also comes with potential biases and errors that can be present in the data. Therefore, it is crucial to provide comprehensive and accurate knowledge during the training phase to mitigate the impact of such biases on the model.
On the other hand, explicit symbolic knowledge requires the existence of corresponding retrieval and query systems. Additionally, the model needs to make judgments on whether to adopt the content of external knowledge. This makes the entire process more complex.

\section{Conclusion}
\label{sec:bibtex}

In this paper, we introduced \name, a framework for generating instruction-tuning data from unsupervised knowledge.
The unsupervised data are used sequentially in the stage of knowledge-guided instruction generation and machine reading comprehension.
Our experiments demonstrate the effectiveness of \name \;in generating diverse, correct, and domain-specific instruction data. By reducing the reliance on human annotators, \name \;offers a promising approach for improving the efficiency and scalability of instruction tuning.

\bibliography{anthology}

\begin{thebibliography}{13}
\expandafter\ifx\csname natexlab\endcsname\relax\def\natexlab#1{#1}\fi

\bibitem[{Joseph et~al.(2016)Joseph, Alber-Morgan, Cullen, and
  Rouse}]{joseph2016effects}
Laurice~M Joseph, Sheila Alber-Morgan, Jennifer Cullen, and Christina Rouse.
  2016.
\newblock The effects of self-questioning on reading comprehension: A
  literature review.
\newblock \emph{Reading \& Writing Quarterly}, 32(2):152--173.

\bibitem[{OpenAI(2022)}]{openaichatgpt}
OpenAI. 2022.
\newblock \href {https://openai.com/blog/chatgpt/} {Chatgpt}.

\bibitem[{OpenAI(2023)}]{openai2023gpt4}
OpenAI. 2023.
\newblock \href {http://arxiv.org/abs/2303.08774} {Gpt-4 technical report}.

\bibitem[{Ouyang et~al.(2022)Ouyang, Wu, Jiang, Almeida, Wainwright, Mishkin,
  Zhang, Agarwal, Slama, Ray, Schulman, Hilton, Kelton, Miller, Simens, Askell,
  Welinder, Christiano, Leike, and Lowe}]{ouyang2022training}
Long Ouyang, Jeff Wu, Xu~Jiang, Diogo Almeida, Carroll~L. Wainwright, Pamela
  Mishkin, Chong Zhang, Sandhini Agarwal, Katarina Slama, Alex Ray, John
  Schulman, Jacob Hilton, Fraser Kelton, Luke Miller, Maddie Simens, Amanda
  Askell, Peter Welinder, Paul Christiano, Jan Leike, and Ryan Lowe. 2022.
\newblock Training language models to follow instructions with human feedback.

\bibitem[{Peng et~al.(2023)Peng, Li, He, Galley, and Gao}]{peng2023instruction}
Baolin Peng, Chunyuan Li, Pengcheng He, Michel Galley, and Jianfeng Gao. 2023.
\newblock Instruction tuning with gpt-4.
\newblock \emph{arXiv preprint arXiv:2304.03277}.

\bibitem[{Radford et~al.(2018)Radford, Narasimhan, Salimans, Sutskever
  et~al.}]{radford2018improving}
Alec Radford, Karthik Narasimhan, Tim Salimans, Ilya Sutskever, et~al. 2018.
\newblock Improving language understanding by generative pre-training.

\bibitem[{Scao et~al.(2022)Scao, Fan, Akiki, Pavlick, Ili{\'c}, Hesslow,
  Castagn{\'e}, Luccioni, Yvon, Gall{\'e} et~al.}]{scao2022bloom}
Teven~Le Scao, Angela Fan, Christopher Akiki, Ellie Pavlick, Suzana Ili{\'c},
  Daniel Hesslow, Roman Castagn{\'e}, Alexandra~Sasha Luccioni, Fran{\c{c}}ois
  Yvon, Matthias Gall{\'e}, et~al. 2022.
\newblock Bloom: A 176b-parameter open-access multilingual language model.
\newblock \emph{arXiv preprint arXiv:2211.05100}.

\bibitem[{Sun et~al.(2023)Sun, Shen, Zhou, Zhang, Chen, Cox, Yang, and
  Gan}]{sun2023principle}
Zhiqing Sun, Yikang Shen, Qinhong Zhou, Hongxin Zhang, Zhenfang Chen, David
  Cox, Yiming Yang, and Chuang Gan. 2023.
\newblock Principle-driven self-alignment of language models from scratch with
  minimal human supervision.
\newblock \emph{arXiv preprint arXiv:2305.03047}.

\bibitem[{Wang et~al.(2022)Wang, Kordi, Mishra, Liu, Smith, Khashabi, and
  Hajishirzi}]{wang2022self}
Yizhong Wang, Yeganeh Kordi, Swaroop Mishra, Alisa Liu, Noah~A Smith, Daniel
  Khashabi, and Hannaneh Hajishirzi. 2022.
\newblock Self-instruct: Aligning language model with self generated
  instructions.
\newblock \emph{arXiv preprint arXiv:2212.10560}.

\bibitem[{Xu et~al.(2023)Xu, Guo, Duan, and McAuley}]{xu2023baize}
Canwen Xu, Daya Guo, Nan Duan, and Julian McAuley. 2023.
\newblock Baize: An open-source chat model with parameter-efficient tuning on
  self-chat data.
\newblock \emph{arXiv preprint arXiv:2304.01196}.

\bibitem[{Zhang(2019)}]{zhang-2019-mc}
Xuanyu Zhang. 2019.
\newblock {MC}{\^{}}2: Multi-perspective convolutional cube for conversational
  machine reading comprehension.
\newblock In \emph{Proceedings of the 57th Annual Meeting of the Association
  for Computational Linguistics}, pages 6185--6190, Florence, Italy.
  Association for Computational Linguistics.

\bibitem[{Zhang and Wang(2020)}]{Zhang2020RceptionWA}
Xuanyu Zhang and Zhichun Wang. 2020.
\newblock Rception: Wide and deep interaction networks for machine reading
  comprehension (student abstract).
\newblock \emph{Proceedings of the AAAI Conference on Artificial Intelligence},
  34(10):13987--13988.

\bibitem[{Zhang et~al.(2022)Zhang, Yang, and Xu}]{zhang-etal-2022-trans}
Xuanyu Zhang, Qing Yang, and Dongliang Xu. 2022.
\newblock {T}ran{S}: Transition-based knowledge graph embedding with synthetic
  relation representation.
\newblock In \emph{Findings of the Association for Computational Linguistics:
  EMNLP 2022}, pages 1202--1208, Abu Dhabi, United Arab Emirates. Association
  for Computational Linguistics.

\end{thebibliography}
\bibliographystyle{acl_natbib}

\end{document}